\documentclass[10pt, letter, onecolumn]{arxiv}

\usepackage{kantlipsum, lipsum}
\usepackage{dm-colors}
\usepackage{amsmath}
\usepackage{pstricks, pst-node}
\usepackage{verbatim}
\usepackage{multirow}
\usepackage{scalerel}
\usepackage{booktabs}
\usepackage{enumitem}
\usepackage{xspace}
\usepackage{bm}
\usepackage{bbm}
\usepackage{mathtools}
\usepackage{soul}
\usepackage{epsfig}
\usepackage{graphicx}
\usepackage{amssymb}
\usepackage{colortbl}
\usepackage{csquotes}
\usepackage{setspace}
\usepackage{colortbl}
\usepackage{tabularx,ragged2e}
\usepackage{placeins}
\usepackage[symbol]{footmisc}
\usepackage[bibstyle=nature,citestyle=numeric-comp,%
            natbib=true,backend=biber,maxbibnames=99,%
            giveninits=false,sorting=none]{biblatex}
\usepackage{nameref}
\usepackage{varioref}
\usepackage[pagebackref=false,breaklinks=false,%
            colorlinks=true,bookmarks=true,citecolor=ourdarkblue,%
            urlcolor=ourdarkblue,linkcolor=ourdarkblue]{hyperref}
\usepackage[noabbrev,capitalize]{cleveref}
\usepackage{etoc}
\usepackage{multirow}
\usepackage{makecell}
\usepackage{tabularx}
\newcolumntype{C}[1]{>{\centering\arraybackslash}m{#1}}

\graphicspath{{figures/}}

\title{\Large{Anomaly Detection in Electrocardiograms: \\Advancing Clinical Diagnosis Through Self-Supervised Learning}}

\author[1,2,$\dagger$]{Aofan Jiang} 
\author[1,2,$\dagger$]{Chaoqin Huang}
\author[3,$\dagger$]{Qing Cao}
\author[3]{Yuchen Xu}
\author[3]{Zi Zeng}
\author[3]{Kang Chen}
\author[1,2,$\ddagger$]{\\Ya Zhang}
\author[1,2,$\ddagger$]{Yanfeng Wang}  

\affil[1]{Shanghai Jiao Tong University, }
\affil[2]{Shanghai Artificial Intelligence Laboratory,\ \ \ \ \ \ \ }
\affil[3]{Ruijin Hospital, Shanghai Jiao Tong University School of Medicine}

\renewcommand{\correspondingauthor}[1]{$\dagger$~Equal contributions: Aofan Jiang, Chaoqin Huang, Qing Cao (\{stillunnamed,huangchaoqin\}@sjtu.edu.cn, cq30553@rjh.com.cn)\\
                                       $\ddagger$~Corresponding authors: Yanfeng Wang and Ya Zhang (\{wangyanfeng622, ya\_zhang\}@sjtu.edu.cn)}

\begin{document}
\begin{refsection}

\begin{abstract}
\textbf{Background:} 
The electrocardiogram (ECG) is an essential tool for diagnosing heart disease, with computer-aided systems improving diagnostic accuracy and reducing healthcare costs. Despite advancements, existing systems often miss rare cardiac anomalies that could be precursors to serious, life-threatening issues or alterations in the cardiac macro/microstructure. We address this gap by focusing on self-supervised anomaly detection (AD), training exclusively on normal ECGs to recognize deviations indicating anomalies.

\textbf{Methods:} 
We introduce a novel self-supervised learning framework for ECG AD, utilizing a vast dataset of normal ECGs to autonomously detect and localize cardiac anomalies. It proposes a novel masking and restoration technique alongside a multi-scale cross-attention module, enhancing the model’s ability to integrate global and local signal features. The framework emphasizes accurate localization of anomalies within ECG signals, ensuring the method’s clinical relevance and reliability. To reduce the impact of individual variability, the approach further incorporates crucial patient-specific information from ECG reports, such as age and gender, thus enabling accurate identification of a broad spectrum of cardiac anomalies, including rare ones.

\textbf{Results:} 
Utilizing an extensive dataset of 478,803 ECG graphic reports from real-world clinical practice, our method has demonstrated exceptional effectiveness in AD across all tested conditions, regardless of their frequency of occurrence, significantly outperforming existing models. It achieved superior performance metrics, including an AUROC of 91.2\%, an F1 score of 83.7\%, a sensitivity rate of 84.2\%, a specificity of 83.0\%, and a precision of 75.6\% with a fixed recall rate of 90\%. It has also demonstrated robust localization capabilities, with an AUROC of 76.5\% and a Dice coefficient of 65.3\% for anomaly localization.

\textbf{Conclusions:} 
The study underscores the potential of AD models in identifying and localizing ECG anomalies, setting a new benchmark for clinical applications. The remarkable performance offers promising avenues for enhancing patient care and diagnostic accuracy in clinical settings.

\end{abstract}

\maketitle


\section{Introduction}
\label{sec:introduction}

In the rapidly evolving landscape of healthcare, the imperative for timely and accurate cardiac condition diagnosis through non-invasive methods has never been more critical~\cite{wearble, ecgsurvey}. The electrocardiogram (ECG), a cornerstone in this diagnostic process, offers a window into the heart's electrical activity, revealing vital insights into cardiac health. Traditional ECG analysis has leaned heavily on recognizing established patterns indicative of specific heart diseases~\cite{wang2023arrhythmia, rahul2022automatic, nankani2022atrial}. However, this approach often overlooks rare or atypical anomalies\footnote{In cases of severe cardiac arrhythmias, like paroxysmal supraventricular tachycardia and ventricular fibrillation, or slower arrhythmias like second-degree type II atrioventricular block and high-degree atrioventricular block, these conditions are often the primary triggers for critical cardiovascular incidents, including shock and cardiogenic sudden death.} that do not conform to well-defined categories, potentially missing critical diagnoses of emerging cardiac conditions. These overlooked anomalies could signal severe, life-threatening conditions or subtle changes in cardiac structure not immediately apparent but equally crucial for patient care~\cite{wuclinical, takayaoutcomes}.

To address this diagnostic gap, building on the foundation anomaly detection (AD)~\cite{chandola2009anomaly, pang2021deep, fernando2021deep, ukil2016iot}, we introduce a novel framework leveraging the intrinsic complexity of ECG signals to detect a wide array of cardiac anomalies, from the most prevalent to the rarest forms, pivoting towards a self-supervised learning approach~\cite{krishnan2022self} that requires no labeled anomaly data. Our model learns from a vast dataset of normal ECGs, enabling it to detect deviations that signify a wide array of cardiac anomalies. This method represents a paradigm shift in ECG analysis, moving from a reliance on manual annotation towards a model that can autonomously identify and localize anomalies, enhancing both the efficiency and accuracy of diagnosis.

Specifically, we propose a novel self-supervised learning framework for ECG AD leveraging both the ECG signal and the ECG report at training. A unique masking and restoration technique is introduced for ECG signal analysis, with its pivotal highlight lies in the integration of a multi-scale cross-attention module to significantly elevates the model's capability to synergize global and local signal features. Such design enables precise identification and localization of subtle anomalies within ECG signals, effectively capturing a broad range of ECG features—from overall signal patterns to the detailed nuances of individual heartbeats—similar to the thorough approach of expert cardiologists~\cite{ecgbook}.

Critically, most existing AD methods in ECG analysis~\cite{li2020survey, venkatesan2018ecg, shen2021time} focus predominantly on analyzing standard ECG signal patterns~\cite{moody2001impact,wagner2020ptb}, frequently overlooking the information available in ECG reports that encapsulates essential details for understanding individual heart conditions~\cite{rasmussenprinterval, alqtinterval}. We tap into this underexploited resource and integrate crucial patient-specific information known to affect the ECG interpretation~\cite{doi:10.1161/CIRCEP.119.007284, moss2010gender} such as age, gender, and other fundamental attributes contained within ECG reports, thus providing a more nuanced understanding of each patient's ECG signal, which significantly enriches the diagnostic process and also effectively reduces the impact of individual variability. This synergy between ECG signal and patient-specific report data not only enhances the accuracy of anomaly detection but also bolsters the model's capacity to navigate the complexities inherent in ECG data, setting a new standard for precision in cardiac diagnostics.

Our approach was rigorously validated using a clinical ECG dataset derived from real hospital settings, underscoring its superiority over various state-of-the-art methods in both identifying and localizing anomalies. A standout feature of our method is its unparalleled capacity to detect a broad spectrum of cardiac anomalies, showcasing remarkable proficiency in pinpointing rare ECG anomalies. This capability highlights the significant practical application of our approach in clinical diagnostics, proving its potential to substantially improve both the accuracy and reliability of cardiac anomaly detection in real-world healthcare environments.

\section{Methods}
\subsection{Dataset}

We conducted a comprehensive assessment by employing an extensive dataset obtained from real-world clinical practice, which includes \textbf{478,803 ECG reports}, each comprising an ECG signal image and a diagnostic summary identifying specific anomalies, spanning from 2012 to 2021. This dataset encapsulates a wide spectrum of ECG abnormalities, ranging from common conditions like atrioventricular block to rare conditions like biventricular hypertrophy, across 102 unique categories. Additionally, the dataset encompasses patient demographics such as gender, age, heart rate, and key cardiac intervals like PR, QT, corrected QT, and QRS complex. We transformed these reports into time-series data for 12-lead ECGs, where each lead represents 2.5 seconds of data, except for lead II, which shows a full 10 seconds, all at a sampling rate of 500Hz.

We gathered all normal ECGs recordings up to the year 2020, accumulating a total of 346,353 cases, for the purpose of model training. The testing phase utilized ECGs from the year 2021, comprising of 132,450 instances of both normal and abnormal ECGs, to assess the model's performance in detecting anomalies across a spectrum of commonality. The abnormal ECGs in the test set were categorized into common, uncommon, and rare based on their frequency of occurrence. Each category contains 34 distinct types of anomalies, totalling to 102 types. These were evenly distributed across three subsets, which contained 60,225, 5,390, and 610 abnormal ECGs, respectively. To create balanced test groups, each subset was paired with an equal number of normal ECGs, forming three distinct test sets: \textbf{common}, \textbf{uncommon}, and \textbf{rare}. Furthermore, to establish a test set with precise anomaly localization annotations at the level of individual signal points, two experienced cardiologists manually mark the exact locations of anomalies on 500 ECGs. Details on the anomaly types across the subsets are available in the appendix.

\begin{figure}[t]
\centering
\includegraphics[width=1\linewidth]{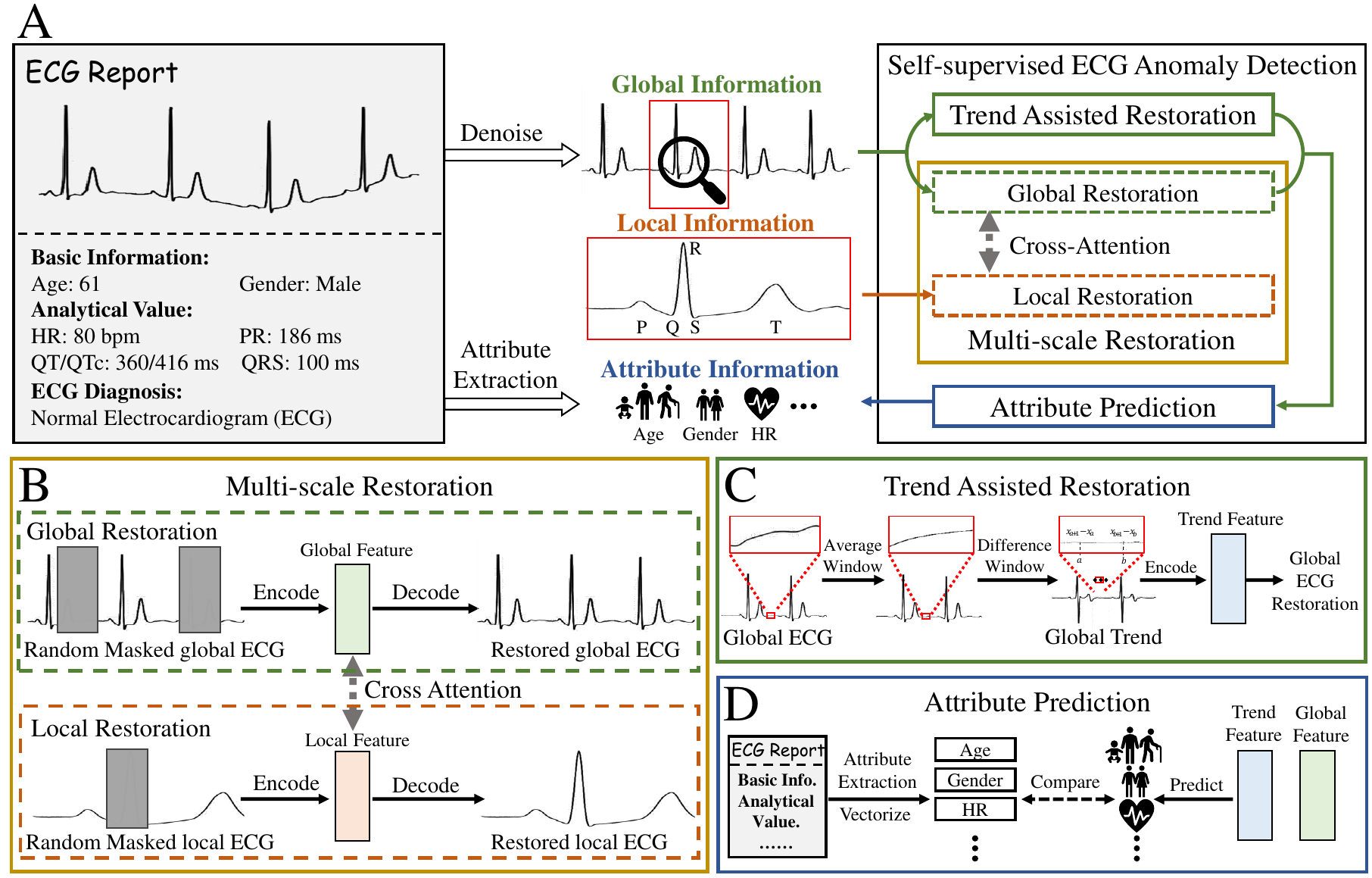}
\caption{The self-supervised ECG anomaly detection framework utilizing both ECG signals and ECG reports.}
\label{fig:model}
\end{figure}

\subsection{Architecture}

Figure~\ref{fig:model} presents a schematic of the novel self-supervised ECG clinical diagnosis framework, specifically designed for the detection and localization of ECG anomalies. This framework utilizes the ECG signal as its input, which is divided into global ECG and local ECG for respective restoration tasks. where the global ECG signal is leveraged to crossly predict attribute information. The framework consists of three main components: (i) multi-scale cross-restoration, (ii) trend assisted restoration, and (iii) attribute prediction module. Detailed explanations of each component are provided in the sections that follow.

\subsubsection{Multi-scale Cross-restoration}

We begin with analyzing a complete global ECG signal, segmenting it into constituent local heartbeats. We randomly select these heartbeats to form a global-local ECG pair for subsequent analysis. This pair is then masked—globally across scattered areas and locally over a specific continuous region. Post-masking, the signals are analyzed through separate global and local encoders, facilitating the extraction of pertinent features from each. Emulating professional cardiologists' approach of evaluating both the ECG's overall context and specific heartbeats details, our model employs a self-attention mechanism inspired by Vaswani et al.~\cite{vaswani2017attention}, which enables the fusion of global and local features into a cohesive cross-attention feature set via concatenation. These integrated features are subsequently dynamically weighted by the self-attention mechanism, directing the model's focus towards signal areas of paramount relevance.

Following this integration, the model proceeds to use two specialized decoders to reconstruct the global and local signals from the extracted features. This step does more than just restore the signals to their original form; it also generates restoration uncertainty maps. These maps visually represent the restoration challenges across different parts of the signal, providing valuable insights into the complexities of the signal restoration task. To refine our model's capability in signal restoration, we introduce an uncertainty-aware restoration loss~\cite{mao2020uncertainty}. The architecture of multi-scale cross-restoration is illustrated in Figure~\ref{fig:model}-B.

\subsubsection{Trend Assisted Restoration}

Figure~\ref{fig:model}-C illustrates how the trend assisted restoration (TAR) method generates a continuous time-series trend by extracting key trend information and reducing signal complexity. It starts by smoothing the global ECG signal with an averaging window during convolution, followed by employing a difference window to identify changes between consecutive points, thereby highlighting the underlying trend. An additional autoencoder, designed for trend analysis, helps restore the global ECG by focusing on this trend information. This approach emphasizes rhythmic patterns over morphological details, enhancing the model's proficiency in restoring and interpreting the global ECG signals.

\subsubsection{Attribute Prediction Module}

To guide the model training, we leverage the rich information from ECG reports, employing a predictive approach to incorporate this data into our framework. As shown in Figure~\ref{fig:model}-D, patient-specific information such as age, gender, and other key attributes is extracted from the reports and transformed into supervisory signals. The attribute prediction module (APM) then uses the global feature and the trend feature extracted from the global restoration and the trend-assisted restoration process, respectively, to predict these attributes, using an auxiliary multi-layer perceptron model. By assessing the model's predictions against the actual attributes using mean square error loss, we enhance the model's capability to discern the impact of these attributes on the ECG diagnosis.

\section{Evaluation}
\label{sec:evaluation}

We employ several evaluation metrics to assess the applicability of our method in real-world clinical scenarios. The performance evaluation for both anomaly detection and localization hinges on the Area Under the Receiver Operating Characteristic Curve (AUROC). To maintain consistency in our assessments across varying contexts, we calculate AUROCs at two levels: patient-level and signal point-level, corresponding to their respective annotations. For anomaly detection, our evaluation is enhanced with additional metrics: sensitivity, to measure the accurate detection of normal cases; specificity, to assess the precise identification of abnormal cases; precision, calculated with a fixed recall rate of 90\%; and the F1 score, to balance precision and recall. In evaluating anomaly localization, we introduce the Dice coefficient, a standard metric for gauging localization precision. Higher values in these metrics indicate a more effective approach. Our methodology is compared against four leading-edge methods: TranAD~\cite{tranad}, AnoTran~\cite{xu2022anomaly}, BeatGAN~\cite{liu2022time} and TSL~\cite{zheng2022task}.

\textbf{Ethics Statement.} The study protocol was approved by the Ruijin Hospital Ethics Committee, Shanghai Jiao Tong University School of Medicine (reference number: 2021-23), and conducted in accordance with the Declaration of Helsinki~\cite{world2013world}.

\begin{table}[t]
 \caption{Anomaly detection and localization results on various evaluation data, compared with state-of-the-art methods. The most effective method is highlighted in \textbf{bold}, and the second-best is \underline{underlined}.
 }
\centering
\begin{tabular}{C{1.7cm}C{2.9cm}C{1.5cm}|C{1.4cm}C{1.4cm}C{1.4cm}C{1.4cm}|C{1.4cm}} 
\toprule
\multirow{2}{*}{Task} & \multirow{2}{*}{Evaluation Data} & \multirow{2}{*}{Metrics} & \multicolumn{3}{r}{State-of-the-art Methods} & & \multirow{2}{*}{Ours}\\
 &  &  & TranAD & AnoTran & BeatGAN  & TSL & \\
\Xcline{1-8}{0.3pt}
\multirow{20}{*}{\makecell[c]{Anomaly \\Detection}} & \multirow{5}{*}{All Test Data} & AUROC  & 0.566  &  0.601   & 0.653   & \underline{0.824} & \textbf{0.912}\\
& &  F1 Score & 0.667  &  0.673   & 0.669  & \underline{0.746} & \textbf{0.837}\\
& & Sensitivity & 0.511  &  0.579   & 0.522  & \underline{0.793} & \textbf{0.842}\\
& & Specificity & 0.514  &  0.564  & \underline{0.698}   & 0.668 & \textbf{0.830}\\
& & Pre@90   & 0.501  &  0.530   & 0.528  & \underline{0.613} & \textbf{0.756} \\
\Xcline{2-8}{0.3pt}
& \multirow{5}{*}{Common Test Set} & AUROC & 0.565  &  0.599   & 0.650   & \underline{0.820} & \textbf{0.914}\\
& & F1 Score  & 0.667  &  0.672   & 0.669   & \underline{0.744} & \textbf{0.839}\\
& & Sensitivity  & 0.512  &  0.576  & 0.537   & \underline{0.792} & \textbf{0.852}\\
& & Specificity  & 0.514  &  0.564   & \underline{0.679}   & 0.662 & \textbf{0.820}\\
& & Pre@90   & 0.501 &  0.529  & 0.528  & \underline{0.611} & \textbf{0.761} \\
\Xcline{2-8}{0.3pt}
& \multirow{5}{*}{Uncommon Test Set} & AUROC  & 0.578  &  0.623   & 0.681   & \underline{0.853} & \textbf{0.895}\\
& & F1 Score  & 0.667 &  0.677   & 0.675   & \underline{0.773} & \textbf{0.819}\\
& & Sensitivity  & 0.504  & 0.628  & 0.552   & \underline{0.753} & \textbf{0.814}\\
& & Specificity  & 0.523 & 0.547   & 0.719   & \underline{0.805} & \textbf{0.828}\\
& & Pre@90   & 0.502  &  0.536   & 0.536   & \underline{0.625} & \textbf{0.700}\\
\Xcline{2-8}{0.3pt}
& \multirow{5}{*}{Rare Test Set} & AUROC  & 0.568  & 0.604   & 0.659  & \underline{0.888} & \textbf{0.896}\\
& & F1 Score  & 0.667  &  0.676  & 0.673   & \underline{0.816} & \textbf{0.827}\\
& & Sensitivity  & 0.513 &  0.582  & 0.510   & \underline{0.769} & \textbf{0.825}\\
& & Specificity  & 0.502  &  0.577   & 0.746  & \textbf{0.885} & \underline{0.831}\\
& & Pre@90   & 0.496  &  0.531   & 0.521  & \textbf{0.693} & \underline{0.692} \\
\hline
\multirow{2}{*}{\makecell[c]{Anomaly \\Localization}} & \multirow{2}{*}{\makecell[c]{Localization\\ Test Set}} & AUROC  & 0.602  & 0.525   & \underline{0.649}  & 0.537 & \textbf{0.765}\\
& & Dice  & 0.549  &  0.511  & \underline{0.580}   & 0.567 & \textbf{0.653}\\
\bottomrule
\end{tabular}
\label{tab:ruijin}
\end{table}

\begin{figure}[h!]
\centering
\includegraphics[width=1.0\textwidth]{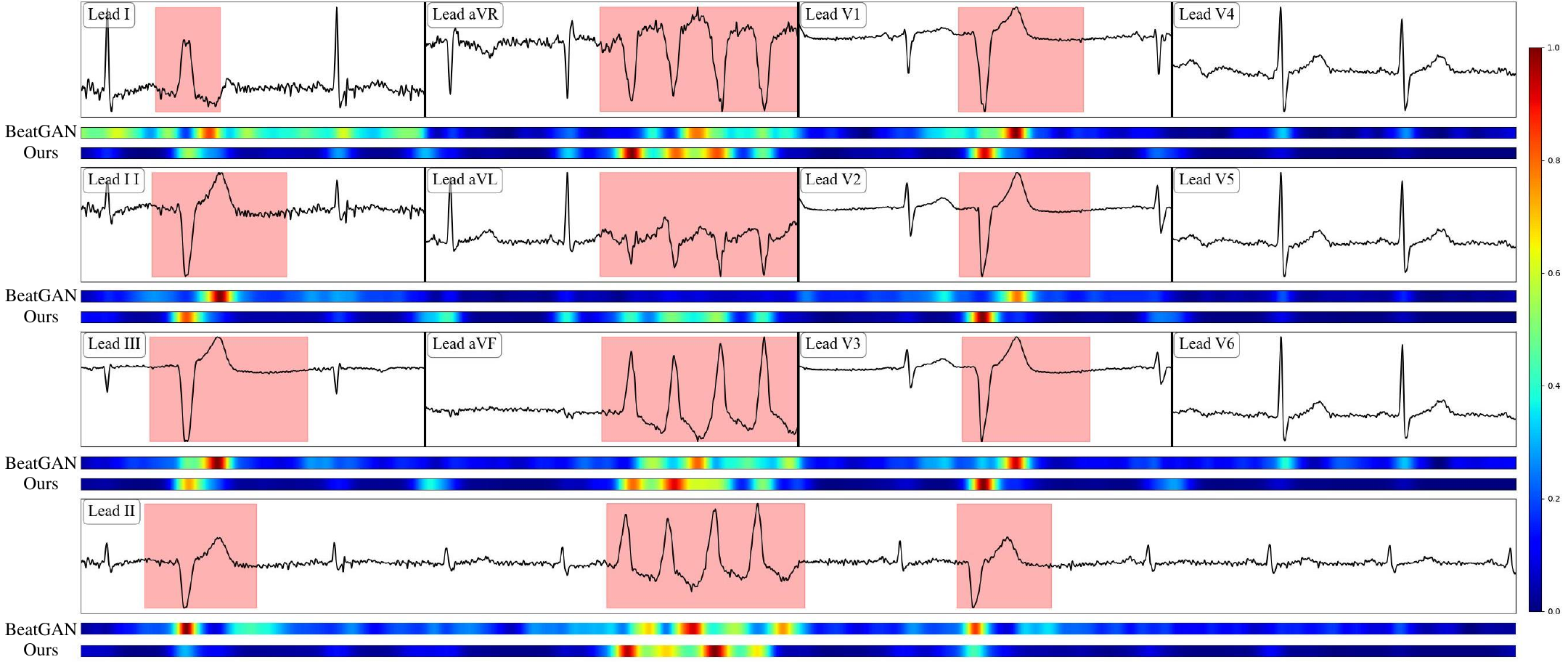}
\caption{Anomaly localization visualization in a 12-lead ECG illustrating ventricular tachycardia and premature ventricular contraction, with ground truth marked by red boxes. Localization results are compared with a leading method, scored between 0 to 1 to indicate anomaly likelihood.}
\label{fig:vis_ruijin}
\end{figure}

\section{Results}
\label{sec:results}

\subsection{Anomaly Detection Performance}

Table~\ref{tab:ruijin} presents the comprehensive evaluation of AD methods on test sets with varying degrees of anomaly rarity. Our method outperforms in most evaluation metrics across all test datasets, achieving a 91.2\% AUROC, 83.7\% F1 score, 84.2\% sensitivity, 83.0\% specificity, and 75.6\% precision at a fixed 90\% recall. These results notably exceed those of competing methods, with gains of 8.8\% in AUROC, 9.1\% in F1 score, and 14.3\% in precision. While methods like TranAD and AnoTran exhibit nearly perfect sensitivity, their specificity scores are lower, indicating challenges in accurately detecting anomalies. Our methodology also demonstrates remarkable effectiveness across varying anomaly rarity levels, with AUROC scores of 91.4\%, 89.5\%, and 89.6\% for the common, uncommon, and rare test sets, respectively, highlighting our method’s effectiveness in detecting rare anomalies.

\subsection{Anomaly Localization Performance}

In the cardiologist-annotated localization test set, our method showcases significant enhancements, achieving the highest results with a 65.3\% Dice coefficient and a 76.5\% AUROC, as detailed in Table~\ref{tab:ruijin}. These results represent a substantial leap forward compared to the closest rival, BeatGAN, with gains of 7.3\% in the Dice score and 11.6\% in AUROC. Methods like TranAD and AnoTran, which do not adequately address individual variances and face challenges in assimilating from large-scale ECG datasets, showed reduced effectiveness in anomaly localization.

For a detailed visual analysis of anomaly localization, Figure~\ref{fig:vis_ruijin} contrasts the performance of our method with that of BeatGAN using a 12-lead ECG test example. The anomaly regions highlighted by the expert cardiologists are in red, serving as the ground truth. Our approach effectively highlights key anomaly areas, such as in Lead aVL, while minimizing false positives in areas considered normal, such as Lead I. 

Our method showcases an acute sensitivity to abrupt changes in signal patterns at a detailed signal point level, leading to more focused anomaly localization than the broader annotations of ground truth. These precise localizations provide deep insights into potential anomalies, offering significant value for assisting medical practitioners. This effectiveness is confirmed by assessments from experienced cardiologists.

\begin{figure}[t]
\centering
\includegraphics[width=0.8\textwidth]{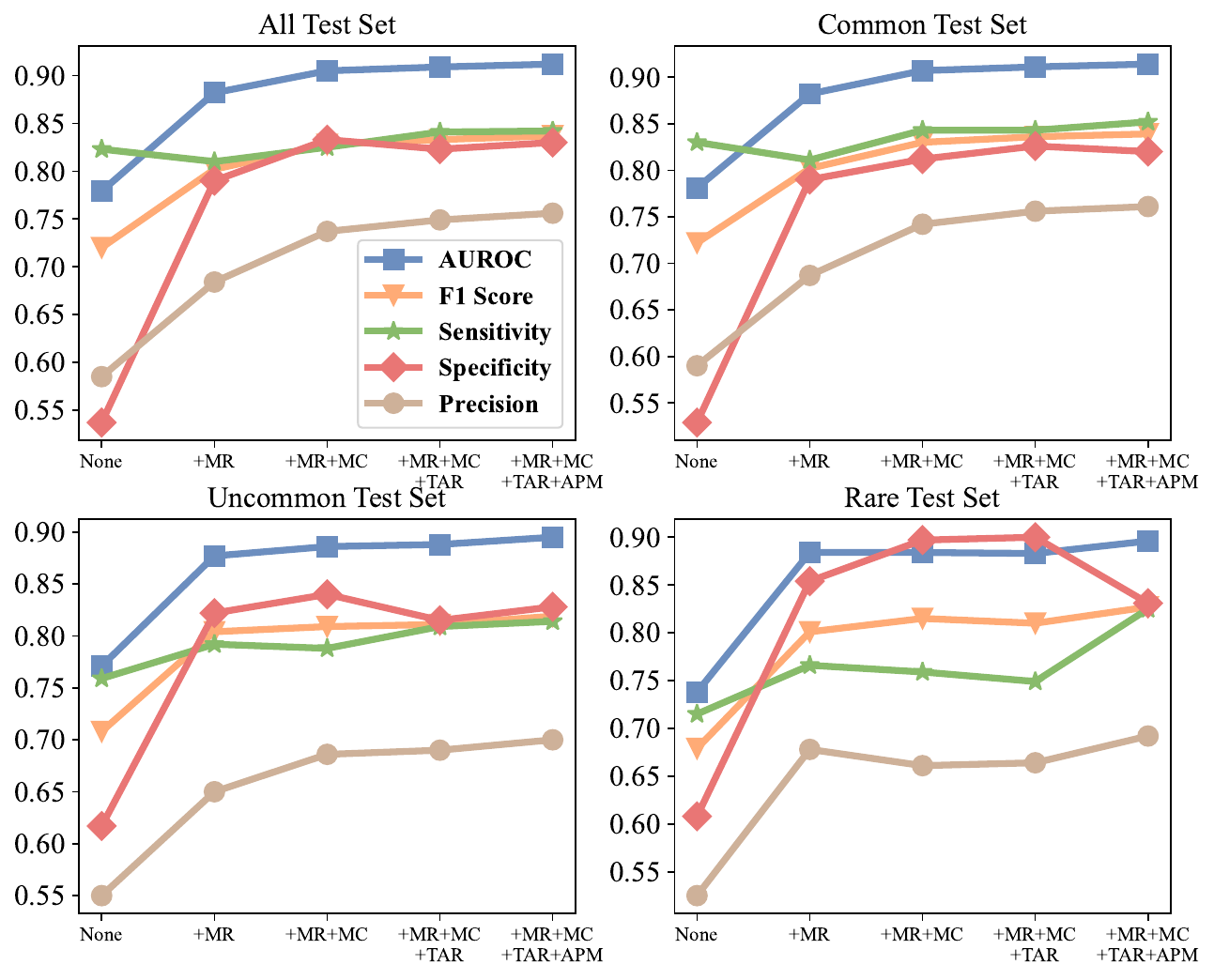}
\caption{Ablation studies evaluating the impact of masking and restoring (MR), multi-scale cross-attention (MC), trend assisted restoration (TAR), and the attribute prediction module (APM) on various test sets.}
\label{fig:abl_ruijin}
\end{figure}

\section{Discussion}

\subsection{Component Analysis}

We perform an ablation study to assess the contribution of each component, starting with a basic model that analyzes global ECG patterns and incrementally introducing additional components: masking and restoring (MR), multi-scale cross-attention (MC), trend assisted restoration (TAR), and the attribute prediction module (APM).

The findings, shown in Figure~\ref{fig:abl_ruijin} and elaborated in the appendix, reveal significant performance improvements with each component's integration. Using the AUROC metric, MR and MC are observed to improve the performance from 78.1\% to 88.2\% and then to 90.7\%, respectively, for the common test set, with similar improvements shown for uncommon and rare test sets. TAR further improved the performance in common and uncommon test sets, albeit with slight performance drop for rare anomalies. Introducing APM leads to the highest AUROC scores across all test sets, achieving peak AUROC scores of 91.4\% for common, 89.5\% for uncommon, and 89.6\% for rare anomalies, demonstrating the holistic advantage of incorporating all modules. Although certain combinations slightly reduced sensitivity/specificity, the complete framework consistently showed superior performance, as evidenced by the top-line scatter points in Figure~\ref{fig:abl_ruijin}, showcasing the positive impact of each module on the overall effectiveness of the system.

\begin{table}[t]
 \caption{Attribute prediction results for various attributes and data types on general hospital ECG dataset. Results are shown in accuracy for binary gender and averaged deviation for other attributes.}
\centering
\scalebox{0.9}{
\setlength{\tabcolsep}{1.3pt}{
\begin{tabular}{C{4.5cm}|C{3.5cm}|C{1.8cm}C{1.8cm}C{1.8cm}C{1.8cm}C{1.8cm}} 
\hline
\multirow{2}{*}{Attribute} & \multirow{2}{*}{Reference Range} & \multirow{2}{*}{Normal} & \multicolumn{4}{c}{Abnormal} \\
& & & All & Common & Uncommon & Rare \\
\hline
Gender & 0 (male) or 1 (female)  & 86.5\% Acc               & 75.9\% Acc           &  76.2\% Acc         & 69.2\% Acc   & 70.9\% Acc \\
Age & 0 $\sim$ 100 years & $\pm$12.6 & $\pm$14.0 & $\pm$13.8 & $\pm$16.8 & $\pm$18.1\\
Heart Rate (HR) & 60 $\sim$ 100 bpm & $\pm$3.64 & $\pm$6.94 & $\pm$6.86 & $\pm$8.20 & $\pm$8.62\\
PR Interval (PR) & 120 $\sim$ 200 ms & $\pm$18.5 & $\pm$26.7 & $\pm$26.2 & $\pm$35.2 & $\pm$37.8\\
QT Interval (QT) & 320 $\sim$ 440 ms & $\pm$20.7 & $\pm$33.0 & $\pm$32.4 & $\pm$43.2 & $\pm$44.0\\
Corrected QT Interval (QTc) & 350 $\sim$ 440 ms & $\pm$21.2 & $\pm$32.7 & $\pm$32.1 & $\pm$42.3 & $\pm$46.4\\
QRS Complex (QRS) & 60 $\sim$ 110 ms & $\pm$5.74 & $\pm$9.76 & $\pm$9.62 & $\pm$12.3 & $\pm$13.6\\
\hline
\end{tabular}
}}
\label{tab:abl_apvalue}
\end{table}

\subsection{Attribute Prediction}

Our study introduces an attribute prediction module (APM) into the ECG analysis framework, aiming to enhance its capability to capture the association between ECG signals and patient attributes. We assessed APM’s effectiveness by examining its accuracy in predicting the attributes across diverse ECG test sets. For binary attributes such as gender, we calculated prediction accuracy, while for numerical attributes, we measured the mean deviation from actual values. The findings, presented in Table 2 and elaborated in the appendix, demonstrate the module’s commendable performance across both normal and abnormal data, exhibiting slightly higher accuracy for normal instances. For instance, gender prediction accuracy decreases from 86.5\% in normal ECGs to 75.9\% in abnormal ECGs. This trend holds numerical attributes as well, where normal ECGs tend to have smaller deviations, notably in age predictions with an average deviation of $\pm12.6$ years for normal ECGs versus $\pm14.0$ years for abnormal ECGs. Moreover, there is a positive correlation between anomaly rarity and attribute prediction accuracy, as exemplified by increasing deviations in heart rate predictions as anomalies become rarer.

\subsection{Clinical Value and Future Work}

Our research incorporates basic heart rhythm analysis akin to “the first glance” of a human cardiologist on electrocardiogram. This approach is particularly relevant as China has recently made significant improvements in its ability to treat cardiovascular diseases through the strategic redistribution of medical resources. However, there remains a pressing need for effective monitoring models in primary healthcare settings. These models must enable non-specialist healthcare providers to accurately assess abnormal ECG signals, regardless of whether they are using low-cost and/or high-end equipment. Such capability is critical for early detection and prevention of acute cardiovascular events, highlighting the importance of our research.

This study emphasizes the clinical utility of the typical 10-second 12-lead ECG for evaluating heart rhythm. However, the limitations of short-term ECGs in identifying intermittent cardiac abnormalities underscore the necessity for extended monitoring. The consensus within the medical community on the need to incorporate long-term ECG data is driven by the potential to uncover novel digital biomarkers and dynamic indicators of cardiovascular health.

Our optimism about expanding our methodology to include more comprehensive datasets stems from its inherent scalability. Looking forward, we aim to develop advanced algorithms capable of processing and integrating diverse types of data. This includes pairing ECG data with echocardiography to create a more nuanced diagnostic tool~\cite{bello2015role}. Such an integrated approach not only promises to enhance diagnostic accuracy but also to identify severe disease conditions requiring prompt medical intervention. By doing so, it aims to ensure that patients do not miss the critical window for diagnosis and treatment, ultimately paving the way for advancements in the prevention and management of cardiovascular diseases.

\section{Conclusion}
\label{sec:conclusion}
This paper introduces a novel self-supervised clinical diagnosis framework for detecting and localizing anomalies in electrocardiograms. Rigorous evaluation on a large-scale ECG dataset obtained from real hospital settings underscores the significant potential of our model for application in real-world clinical diagnosis scenarios.

\setlength\bibitemsep{5pt}
\printbibliography
\balance
\clearpage

\newpage
\clearpage
\onecolumn

\renewcommand{\thesection}{A.\arabic{section}}
\renewcommand{\thefigure}{A.\arabic{figure}}
\renewcommand{\thetable}{A.\arabic{table}} 
\renewcommand{\theequation}{A.\arabic{equation}} 
\renewcommand{\theHsection}{A\arabic{section}}

\setcounter{section}{0}
\setcounter{figure}{0}
\setcounter{table}{0}
\setcounter{equation}{0}

\noindent \textbf{\LARGE{Appendix}}\\
\normalfont

In the following sections, we report related work, modelling details, additional experiments and analysis to further illustrate the performance of the proposed ECG anomaly detection model. We provide details on:

\begin{itemize}
\item  Related Works
\item  Framework Details
\item  Model Training and Evaluation Procedure
\item  ECG Dataset Details
\item  Additional Results Tables and Figures
\item  Additional Experiments on Publicly Available ECG Datasets
\begin{itemize}
    \item Performance Analysis on PTB-XL Dataset
    \item Performance Analysis on MIT-BIH Dataset
\end{itemize}
\end{itemize}

\section{Related Works}

\subsection{Anomaly Detection}

The rarity and variation of abnormal samples, coupled with the abundance of normal instances in data, makes it challenging to obtain a sufficient number of labeled anomalous instances for training. Due to the limited number of labeled anomalous samples available for direct training, anomaly detection tasks are typically approached as unsupervised learning tasks. As a result, anomaly detection often relies on utilizing normal samples for training and identifying any samples that deviate from this norm as anomalies. The objective of object anomaly detection is to effectively distinguish between normal and anomalous samples, which can be viewed as a binary classification problem. In contrast to traditional machine learning techniques such as one-class support vector machines~\cite{scholkopf2001estimating} and support vector data description~\cite{tax2004support}, contemporary deep learning methods can be classified into three categories: one-class discriminative approaches, reconstruction-based approaches and self-supervised learning-based approaches.

One-class discriminative methods aim to learn a decision boundary that distinguishes between normal and abnormal samples. Chalapathy \textit{et al.}~\cite{chalapathy2018ocad1} utilize a robust autoencoder model to mitigate the constraints inherent in Principal Component Analysis (PCA) and robust PCA, achieving this by learning a nonlinear subspace capable of capturing the majority of data points while accommodating arbitrary corruption. Ruff \textit{et al.}~\cite{ruff2018ocad2} enhance classical support vector data description~\cite{tax2004support} by training a neural network to map data into a hypersphere of minimum volume, effectively extracting common factors of normal variation. Zong \textit{et al.}~\cite{zong2018deep} apply the Gaussian Mixture Model to the low-dimensional representation produced by the autoencoder, facilitating the direct estimation of mixture membership for individual samples.

Reconstruction-based methods~\cite{guo2023encoder, chen2021mama, 9469869, zhang2019deep, audibert2020usad, li2019mad} employ generative neural networks to reconstruct normal samples, operating under the assumption that a model trained on normal samples will struggle to accurately reconstruct abnormal regions. Zhang \textit{et al.}~\cite{zhang2019deep} employ a convolutional autoencoder and long-short-term-memory network to reconstruct multi-scale signature matrices for the purpose of detecting multivariate time series anomalies. Audibert \textit{et al.}~\cite{audibert2020usad} introduce adversarial training into the reconstruction framework, aiming to enhance the model's ability to amplify the reconstruction error of inputs containing anomalies, thereby achieving increased stability. Li \textit{et al.}~\cite{li2019mad} employ generative adversarial networks with long-short-term-memory recurrent neural networks as base models, with the primary objective of effectively detecting anomalies by leveraging spatial-temporal correlations and dependencies among multiple variables.

Self-supervised learning-based methods explore the use of proxy tasks to improve the representation learning of normal samples. Li \textit{et al.}~\cite{li2023self} employ both multi-scale cropping and a novel augmentation task called CropMixPaste to synthesize anomaly samples, along with a simple masked attentive predicting block for refining local patterns. Dong \textit{et al.}~\cite{dong2023swssl} utilize sliding window-based self-supervised learning to allow the network to process patches instead of resizing images to low resolution, applying the Barlow Twins loss and Continuity Preserving loss for augmentation-invariant feature learning. In this study, our focus is on developing a novel self-supervised proxy task for detecting anomalies in Electrocardiograms.

\subsection{Computer-aided ECG Diagnosis}

The conventional intelligent algorithm employed for ECG diagnosis typically comprises three principal stages: data preprocessing, feature extraction, and classification~\cite{liu2021deep}. The advancement of deep learning has greatly enhanced the system's ability to extract features, resulting in a higher accuracy in classifying ECG signals. However, current researches following classification framework require a large amount of labeled abnormal data and are limited to detecting only the specific anomaly types provided in the training data, such as arrhythmia classification~\cite{wang2023arrhythmia, rahul2022automatic}, atrial fibrillation classification~\cite{nankani2022atrial, feng2022novel}, or multi-label classification~\cite{du2021fm, cao2021practical}, thereby falling short of identifying all unseen abnormal ECG signals. This constraint diminishes their practical utility in clinical diagnosis. In contrast, anomaly detection has the potential to identify all possible anomalies distinct from the modeled normal ECG features, offering a more comprehensive approach to clinical diagnosis.

\subsection{Anomaly Detection in Electrocardiogram}

Current research frames ECG anomaly detection within the broader scope of time-series anomaly detection, primarily focusing on two key approaches: reconstruction-based methods~\cite{xu2022anomaly, tranad, zhou2019beatgan} and self-supervised learning-based methods~\cite{zheng2022task}. AnoTran~\cite{xu2022anomaly} and TranAD~\cite{tranad} leverage the transformer network architecture~\cite{vaswani2017attention} for ECG signal processing and reconstruction, facilitating the identification of normal ECG patterns to detect anomalies effectively. Specifically, AnoTran~\cite{xu2022anomaly} introduces an association discrepancy measure to capture the associations and temporal dependencies within time series data while TranAD~\cite{tranad} employs a reconstruction loss function that encourages the model to learn a robust representation of the normal patterns in the time series data. Despite their innovations, these strategies do not account for individual variability, limiting their applicability in clinical diagnostics. BeatGAN~\cite{zhou2019beatgan} utilizes a reconstruction-based framework and employs a generative adversarial network~\cite{goodfellow2020generative} to reconstruct normal heartbeats rather than the complete raw ECG signal. By comparing the residuals between the input heartbeat and the reconstructed heartbeat, significant differences can be identified as anomalies in individual heartbeats. While this approach considers individual differences at the heartbeat level, it neglects the global rhythm information that is crucial for accurate clinical diagnosis. TSL~\cite{zheng2022task} adopt a self-supervised learning approach, aiming to detect abnormalities at a patient level across the entire ECG signals. Originally devised for EEG signal anomalies, this method uses a 3-class convolutional neural network trained on artificially generated abnormal ECG data, which varies in waveform amplitude and frequency among individuals. Although innovative, the reliance on simulated anomalies raises concerns about the method’s ability to accurately capture the complex and diverse nature of real-world ECG abnormalities, underscoring the necessity for a more appropriate self-supervised learning task.

\section{Framework Details}

\begin{figure}[t]
\centering
\includegraphics[width=1.0\textwidth]{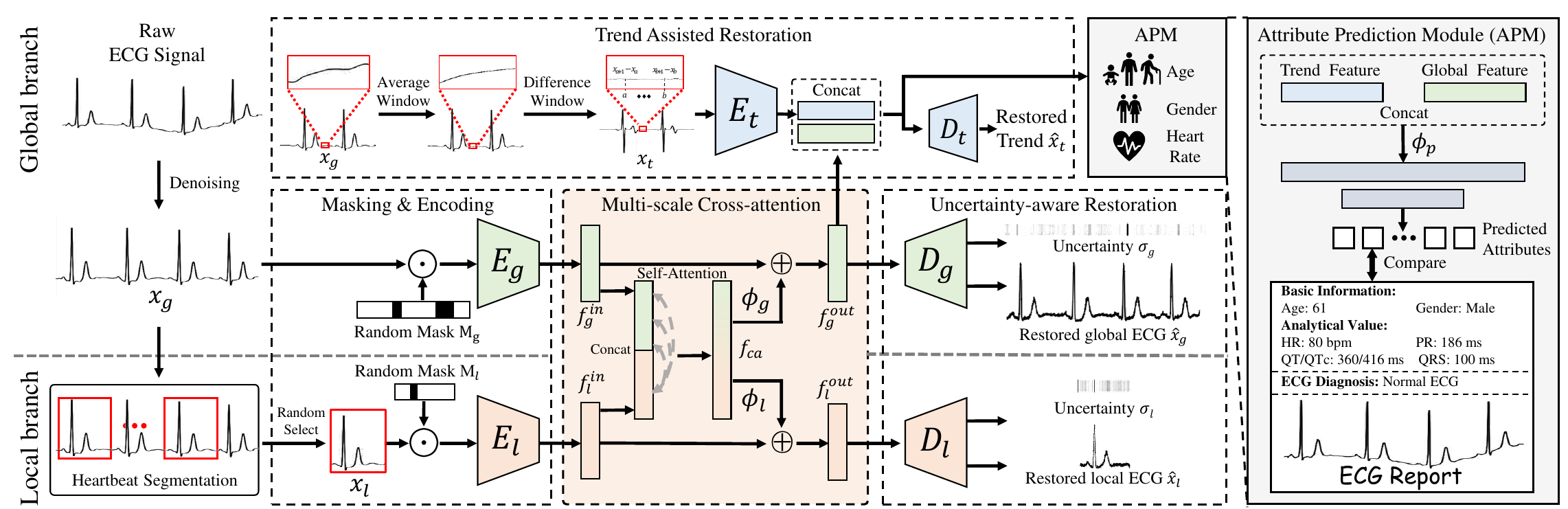}
\caption{The details of multi-scale cross-restoration framework for ECG anomaly detection.}
\label{fig:model_supp}
\end{figure}

We concentrate on the detection and precise localization of anomalies in Electrocardiograms (ECGs). Our strategy is distinctive in that it relies solely on normal ECG data for training the model, aiming to recognize various abnormal ECG patterns and accurately identify specific anomalous regions within ECGs. 

To clearly define our method, we utilize a training dataset represented as $\{x_i, i=1, ..., N\}$, where each $x_i$ is the $i$-th normal ECG signal, consisting of $D$ signal points. Our model is engineered to calculate anomaly scores, expressed as $A(x_{test})$, and to produce score maps, denoted as $S(x_{test})$, for any new test ECG, $x_{test}$.

In Supplementary Figure~\ref{fig:model_supp}, we present a detailed schematic of the self-supervised, multi-scale cross-restoration framework, specifically designed for the detection and localization of ECG anomalies. The framework consists of three main components: (i) multi-scale cross-restoration, (ii) trend assisted restoration, and (iii) attribute prediction module, elaborately expounded upon in subsequent sections.

\subsection{Multi-scale Cross-restoration}

Given the denoised global ECG signal $x_{g}\in \mathbb{R}^D$, we segment the signal into individual heartbeats and stochastically select one heartbeat $x_{l}\in\mathbb{R}^d$ to constitute the global and local ECG pair, serving as input to the framework. Subsequently, two masks are applied to the input pair, with the value of the masked region set to zero. Specifically, the global mask $M_g$ randomly masks multiple dispersed regions across the entire global ECG, while the local mask $M_l$ randomly masks a consecutive region on the local heartbeat. The ultimate global feature $f^{in}_g$ and local feature $f^{in}_l$ are derived from the masked signals, obtained separately through global encoder $E_g$ and local encoder $E_l$. This process is mathematically expressed as:
\begin{equation}
\centering
f^{in}_g = E_g (x_{g}\odot M_g),\quad f^{in}_l = E_l (x_{l}\odot M_l),
\end{equation}
where $\odot$ represents the element-wise product.

As delineated in the primary document, we employ the self-attention mechanism~\cite{vaswani2017attention} on global feature $f^{in}_g$ and local feature $f^{in}_l$. The well-established self-attention mechanism operates with three identical terms, namely Query (Q), Key (K), and Value (V), as input. Adhering to this mechanism, we derive the cross-attention feature denoted as $f_{ca}$ by setting these three items equal to the concatenation of the global and local features. This process is mathematically expressed as follows:
\begin{equation}
    f_{ca} = Attention(Q,K,V)=\mathrm{softmax}(\frac{QK^T}{\sqrt{d_k}})V,\quad  Q=K=V=concat(f^{in}_g,f^{in}_l),
\end{equation}
where $\sqrt{d_k}$ serves as a scaling factor, representing the square root of the feature dimension. Subsequent to the dynamic assignment of weights to the concatenated features by the self-attention mechanism, two Multi-Layer Perceptron (MLP) architectures $\phi_g (\cdot)$ and $\phi_l (\cdot)$ are applied to the cross-attentioned feature. Specifically, the output features $f^{out}_g$ and $f^{out}_l$ containing cross-scale information are obtained through the residual connection, \textit{i.e.},
\begin{equation}
f^{out}_g = f^{in}_g + \phi_g(f_{ca}),\quad f^{out}_l = f^{in}_l + \phi_l(f_{ca}).
\end{equation}

With the objective of signal restoration, the features encoded in $f^{out}_g$ and $f^{out}_l$ undergo decoding by two distinct decoders, denoted as $D_g$ and $D_l$, resulting in the restored signals $\hat{x}_g$ and $\hat{x}_l$, respectively. Concurrently, corresponding restoration uncertainty maps $\sigma_g$ and $\sigma_l$ are generated, quantifying the level of difficulty associated with restoring different signal points. This process is formally expressed as $\hat{x}_g, \sigma_g = D_g(f^{out}_g)$ and $\hat{x}_l, \sigma_l = D_l(f^{out}_l)$. To integrate the uncertainty related to restoration into loss function, we employ an uncertainty-aware restoration loss:
\begin{equation}
\mathcal{L}_{global}=\sum_{k=1}^D \left\{\frac{(x_{g}^{k}-\hat{x}_{g}^{k})^2}{\sigma^{k}_{g}}+\log \sigma^{k}_{g}\right\},
\quad
\mathcal{L}_{local}=\sum_{k=1}^d \left\{\frac{(x_{l}^{k}-\hat{x}_{l}^{k})^2}{\sigma^{k}_l}+\log \sigma^{k}_{l}\right\}.
\end{equation}  
Specifically, each loss function involves normalizing the first term by its respective uncertainty, while the second term is introduced to prevent the prediction of a large uncertainty for all restoration signal points, in accordance with the approach proposed by Mao et al.~\cite{mao2020uncertainty}. Here, the superscript $k$ designates the location of the $k$-th element within the signal. It is noteworthy that, in contrast to~\cite{mao2020uncertainty}, the application of the uncertainty-aware loss is directed towards the restoration process rather than the reconstruction.

\subsection{Trend Assisted Restoration}

The trend assisted restoration (TAR), as depicted in Supplementary Figure~\ref{fig:model_supp}, is designed to generate a continuous time-series trend denoted as $x_t \in \mathbb{R}^D$. Another autoencoder tailored for processing trend information is introduced to assist in restoring the masked global ECG based on the complete trend signal. The restored signal $\hat{x}_t$ is decoded from the combination of trend feature and global ECG feature. This process can be expressed mathematically $\hat{x}_t= D_t(\text{concat}(E_t(x_{t}),f^{out}_g))$, where $E_t$ and $D_t$ represent the trend encoder and decoder. After obtaining the restored global signal, the restoration loss is intuitively calculated using the Euclidean distance, \textit{i.e.},
\begin{equation}
\mathcal{L}_{trend}=\sum_{k=1}^D (x_{g}^{k}-\hat{x}_{t}^{k})^2,  
\end{equation}
where the superscript k denotes the $k$-th signal point in global ECG. 

\subsection{Attribute Prediction Module}

In addition to the ECG signal, we extract attribute information embedded within textual reports for comparison with the outcomes generated by the attribute prediction module. Assuming $m$ distinct attributes, we normalize and vectorize these values into a vector $Y:=[y_1, ... ,y_m]$, with each $y_i$ ranging between [0,1]. We use the global pattern feature from the trend generation module to predict these attributes relative to the ECG's overall state. This prediction uses an additional MLP $\phi_p(\cdot)$, expressed as $\hat{Y} = \phi_p(\text{concat}(E_t(x_{t}),f^{out}_g))$, where $\hat{Y}:=[\hat{y}_1, ... ,\hat{y}_m]$. We then calculate a mean-square-error loss between the predicted and actual attributes as follows:
\begin{equation}
    \mathcal{L}_{pred}(Y,\hat{Y}) = \frac{1}{m}\sum_{i=1}^m (y_i-\hat{y}_i)^2.
\end{equation}

\subsection{Implementation Details}

The ECG undergoes pre-processing through a Butterworth filter and Notch filter~\cite{van2019heartpy} to eliminate high-frequency noise and mitigate ECG baseline wander. R-peaks are identified utilizing an adaptive threshold methodology outlined in paper~\cite{van2019rpeak}, which does not necessitate the incorporation of learnable parameters. Subsequently, the positions of the identified R-peaks are employed to segment the ECG sequence, delineating a series of individual heartbeats.

We employ a convolutional-based autoencoder architecture, as outlined in paper~\cite{liu2022time}. The model undergoes training using the AdamW optimizer, initializing with a learning rate of 1e-4 and a weight decay coefficient of 1e-5, over a span of 50 epochs on a singular NVIDIA GTX 3090 GPU. Decay scheduling is implemented through a single cycle of cosine learning rate. The batch size is configured to 32.

\section{Model Training and Evaluation Procedure}
\subsection{Loss Function}

During the training phase, the comprehensive loss function for optimization is formulated as:
\begin{equation}
\mathcal{L} = \mathcal{L}_{global} + \alpha \mathcal{L}_{local}+\beta \mathcal{L}_{trend} + \gamma \mathcal{L}_{pred}, 
\end{equation}
where $\alpha$, $\beta$, and $\gamma$ serve as trade-off parameters, weighting the contributions of the individual loss components. For simplicity, we adopt the default values $\alpha = \beta = \gamma = 1.0$.

\subsection{Evaluation}
For each ECG, $x_{test}$, in the testing phase, local ECGs are iteratively selected from the segmented heartbeats $\{x_{l,m},m=1,...,M\}$ instead of being randomly chosen as in the training phase. The selected local ECG is then combined with the global test ECG to form the input pair. 

The score map $S(x_{test})$ results from the combination of predicted outcomes from both the global and local branches, denoted as $S_g(x_{test})$ and $S_l(x_{test})$ respectively. Mathematically, this combination is expressed as $S(x_{test}) = S_g(x_{test}) + S_l(x_{test})$.
Here, the predicted score map $S_g(x_{test})$ for the global branch is given by:
\begin{equation}
S_g(x_{test}) =  \frac{(x_{test}-\hat{x}_{test})^2}{\sigma_g} + (x_{test}-\hat{x}_{t})^2.
\end{equation}

For the local branch, the predicted score map is obtained by summing the score map of each heartbeat with its corresponding original position in the global ECG, formulated as:
\begin{equation}
S_l(x_{test}) = \sum_{m=1}^M I_m \odot \frac{(x_{l,m}-\hat{x}_{l,m})^2}{\sigma_{l,m}}, 
\end{equation}
where $I_m \in \mathbb{R}^D$ is a zero-one indicator vector, assuming a value of one solely at the position of the $m$-th segmented heartbeat.

The anomaly score $A(x_{test})$ for the whole image is defined as the mean value of each signal point in the score map $S(x_{test})$. Both in the score map and anomaly score, higher values are indicative of a higher likelihood of anomalies.

\begin{table}[h!]
\caption{Detailed information on our general hospital ECG dataset.}
\centering
\scalebox{0.9}{
\begin{tabular}{C{1.5cm}|C{1.2cm}C{1.2cm}C{1.0cm}C{10.0cm}} 
\hline
\multirow{2}{*}{Data} & \multicolumn{2}{c}{\#ECGs} & \multirow{2}{*}{\#Anomaly Types} & \multirow{2}{*}{Anomaly Types}\\
& Normal & Abnormal & & \\
\hline
Train & 346,353 & 0 & 1 & Normal ECG\\
\hline
Test (all) & 66,225 & 66,225 & 102 & All below\\
\hline
\multicolumn{1}{c|}{Test (Common)} & 60,225 & 60,225 & 34 & \multicolumn{1}{p{10cm}}{T-wave changes, Sinus bradycardia, ST-T segment changes, Sinus tachycardia, ST segment changes, Atrioventricular (AV) block, Left ventricular hypertrophy, Atrial fibrillation, Right bundle branch block, Atrial premature beat, Sinus arrhythmia, Ventricular premature beat, Low voltage, Left anterior fascicular block, ST segment elevation, Incomplete right bundle branch block, ST segment depression, T-wave peaked, Clockwise rotation, Atrial flutter, Prominent U-wave, Complete left bundle branch block, Ventricular pacing rate, Atrial tachycardia, Intraventricular block, Left ventricular hypertrophy, Paroxysmal atrial tachycardia, Inferior myocardial infarction, Frequent atrial premature beats, Extensive anterior wall myocardial infarction, Anterior myocardial infarction, Frequent ventricular premature beats, Abnormal Q waves, Prolonged QT interval.}\\
\hline
\multicolumn{1}{c|}{Test (Uncommon)} & 5,390 & 5,390 & 34 & 
\multicolumn{1}{p{10cm}}{Right axis deviation, Old anterior wall myocardial infarction, Old inferior wall myocardial infarction, Left axis deviation, Atrial pacing rhythm, Preexcitation syndrome, Interventricular conduction delay, Low voltage T wave, Poor R wave progression or reversed R wave progression, Atrial premature beat not conducted, Junctional escape beat, Ventricular premature beat interpolated, Short PR interval with normal QRS complex, Second degree atrioventricular block, Runs of atrial premature beats, Couplets of atrial premature beats, Paroxysmal supraventricular tachycardia, J point elevation, Paced electrocardiogram, Tall peaked P waves, Runs of ventricular premature beats, Non-paroxysmal junctional tachycardia, Bifid P wave, Third degree atrioventricular block, Mobitz type I second degree atrioventricular block, Horizontal ST segment depression, Junctional premature beat, Ventricular tachycardia, Ventricular escape beat, Anterior wall myocardial infarction, T wave inversion, Posterior wall myocardial infarction, Long RR interval, Triplet of atrial premature beats.}\\
\hline
\multicolumn{1}{c|}{Test (Rare)} & 610 & 610 & 34 & \multicolumn{1}{p{10cm}}{Short run of ventricular tachycardia, High lateral wall myocardial infarction, upsloping ST segment depression, downsloping ST segment depression, second-degree sinoatrial block, high-degree atrioventricular block, right ventricular infarction, dextrocardia, Right ventricular hypertrophy, couplet ventricular premature contractions, second-degree type II sinoatrial block, second-degree type I sinoatrial block, atrial tachycardia with variable AV block, sinus arrest, bidirectional P waves, lateral wall myocardial infarction, old lateral wall myocardial infarction, second-degree type II atrioventricular block, Left posterior fascicular block, bidirectional T waves, old posterior wall myocardial infarction, inverted P waves, wandering atrial pacemaker, triplet ventricular premature contractions, subendocardial myocardial infarction, Old high lateral wall myocardial infarction, atrial escape rhythm, atrial fibrillation, widened P waves, atrial couplet, ventricular fibrillation, shortened QT interval, alternating left and right bundle branch block, biventricular hypertrophy.}\\
\hline
\end{tabular}
}
\label{tab:dataset}
\end{table}

\begin{table}[h!]
 \caption{Ablation studies on general hospital ECG dataset. Factors under analysis are: the masking and restoring (MR), the multi-scale cross-attention (MC), trend assisted restoration (TAR), and the attribute prediction module (APM). Results are shown in the patient-level AUC. The best-performing method is in \textbf{bold}.
 }
\centering
\scalebox{0.9}{
\begin{tabular}{C{3.0cm}C{1.5cm}|C{1.2cm}C{1.2cm}C{1.5cm}C{2.6cm}|C{3.5cm}} 
\toprule
Evaluation Data & Metrics & None & +MR & +MR+MC  & +MR+MC+TAR & +MR+MC+TAR+APM\\
\Xcline{1-7}{0.3pt}
\multirow{5}{*}{All Test Data} & AUROC  & 0.779  &  0.882   & 0.905   & 0.909 & \textbf{0.912}\\
&  F1 Score & 0.720  &  0.802   & 0.828  & 0.833 & \textbf{0.837}\\
& Sensitivity & 0.823  &  0.810  & 0.825  & 0.841 & \textbf{0.842}\\
& Specificity & 0.537  &  0.790 & \textbf{0.833}   & 0.823 & 0.830\\
 & Pre@90   & 0.585  &  0.684   & 0.737  & 0.749 & \textbf{0.756} \\
\Xcline{1-7}{0.3pt}
 \multirow{5}{*}{Common Test Set} & AUROC & 0.781  &  0.882   & 0.907   & 0.911 & \textbf{0.914}\\
 & F1 Score  & 0.722  &  0.802   & 0.830   & 0.836 & \textbf{0.839}\\
 & Sensitivity  & 0.830  &  0.811  & 0.843   & 0.843 & \textbf{0.852}\\
 & Specificity  & 0.529  &  0.790   & 0.812   & \textbf{0.826} & 0.820\\
 & Pre@90   & 0.590 &  0.687  & 0.742  & 0.756 & \textbf{0.761} \\
\Xcline{1-7}{0.3pt}
 \multirow{5}{*}{Uncommon Test Set} & AUROC  & 0.771  &  0.877   & 0.886   & 0.888 & \textbf{0.895}\\
 & F1 Score  & 0.708 &  0.804   & 0.809   & 0.811 & \textbf{0.819}\\
 & Sensitivity  & 0.759  & 0.792  & 0.788   & 0.809 & \textbf{0.814}\\
 & Specificity  & 0.617 & 0.822   & \textbf{0.840}   & 0.815 & 0.828\\
 & Pre@90   & 0.550  &  0.650   & 0.686   & 0.690 & \textbf{0.700}\\
\Xcline{1-7}{0.3pt}
 \multirow{5}{*}{Rare Test Set} & AUROC  & 0.738  & 0.884   & 0.884  & 0.883 & \textbf{0.896}\\
 & F1 Score  & 0.679  &  0.801  & 0.815   & 0.810 & \textbf{0.827}\\
 & Sensitivity  & 0.715  &  0.766  & 0.759   & 0.749 & \textbf{0.825}\\
 & Specificity  & 0.608  &  0.854   & 0.897  & \textbf{0.900} & 0.831\\
& Pre@90   & 0.525  &  0.678   & 0.661  & 0.664 & \textbf{0.692} \\
\bottomrule
\end{tabular}}
\label{tab:ruijin_detail}
\end{table}

\subsection{Evaluation Protocols}

The evaluation metrics employed in our paper include patient/signal point level AUC (area under the Receiver Operating Characteristic curve), F1 score, sensitivity, specificity, precision (@recall=90\%), and dice.

AUC is a widely used metric to evaluate the performance of binary classification models. It quantifies the ability of a model to discriminate between positive and negative classes across various threshold values. A higher AUC value indicates better discrimination capability, with a value of 1 representing perfect classification and 0.5 representing random guessing.

Sensitivity, also known as true positive rate or recall, measures the proportion of actual positive cases that are correctly identified by the model. It is calculated as
$Sensitivity =\frac{TP}{TP+FN}$. Higher sensitivity indicates that the model effectively captures positive cases.

Specificity measures the proportion of actual negative cases that are correctly identified by the model. It is calculated as $Specificity = \frac{TN}{TN+FP}$. Higher specificity indicates that the model effectively excludes negative cases.

Precision, also known as positive predictive value, represents the proportion of correctly predicted positive cases out of all instances predicted as positive, $Precision = \frac{TP}{TP+FP}$. Precision calculated at a specific recall level (in this case, at a recall of 90\%) provides insight into the model's performance when a higher recall threshold is desired.

The F1 score is a measure of a model's accuracy that considers both precision and recall. It is the harmonic mean of precision and recall, calculated as $F1 = \frac{2\times Precision \times Recall}{Precision + Recall}$. F1 score ranges from 0 to 1, with higher values indicating better model performance in terms of both precision and recall.

The Dice coefficient is a measure of similarity between two sets, quantifying the overlap between the predicted and ground truth anomaly masks. It is calculated as $Dice = 2\times \frac{Intersection}{Union}$, where the intersection represents the overlapping area between the predicted and ground truth masks, and the union represents the total area encompassed by both masks. For these evaluation protocols, $TP, TN, FP, FN$ represent true positives, true negatives, false positives, and false negatives, respectively.

\section{ECG Dataset Details}

\subsection{Open Source Datasets}

\textbf{PTB-XL}~\cite{wagner2020ptb} database includes clinical 12-lead ECGs lasting for 10 seconds with a sampling rate of 500Hz for each patient, with annotations only at the patient-level. In order to create a new and challenging benchmark for anomaly detection and localization, 8167 normal ECGs were used for training, while 912 normal and 1248 abnormal ECGs were used for testing. We provide signal point-level annotations for 400 ECGs, including 22 different types of abnormalities, which were annotated by two experienced cardiologists. To the best of our knowledge, we are the first to explore ECG anomaly detection and localization across various patients on such a complex and large-scale publicly available database.

\noindent\textbf{MIT-BIH}~\cite{moody2001impact} arrhythmia dataset partitions the ECGs from 44 patients into individual heartbeats, delineated by the annotated R-peak position. This segmentation is conducted following the approach outlined in paper~\cite{liu2022time}. For training purposes, a total of 62,436 normal heartbeats are employed, while testing utilizes 17,343 normal heartbeats and 9,764 abnormal heartbeats. The annotations at the heartbeat level signify whether each heartbeat is categorized as normal or abnormal.

\subsection{Clinical Public Hospital ECG Dataset}

The comprehensive details of the general hospital ECG dataset, including the specific anomaly types present in each test set, are enumerated in Supplementary Table~\ref{tab:dataset}. The training data comprises solely one anomaly type, specifically normal ECG, whereas the test data encompasses a total of 102 anomaly types except normal ECG, with each subset of the test set featuring 34 distinct anomaly types individually.

\begin{figure}[h!]
\centering
\includegraphics[width=1.0\textwidth]{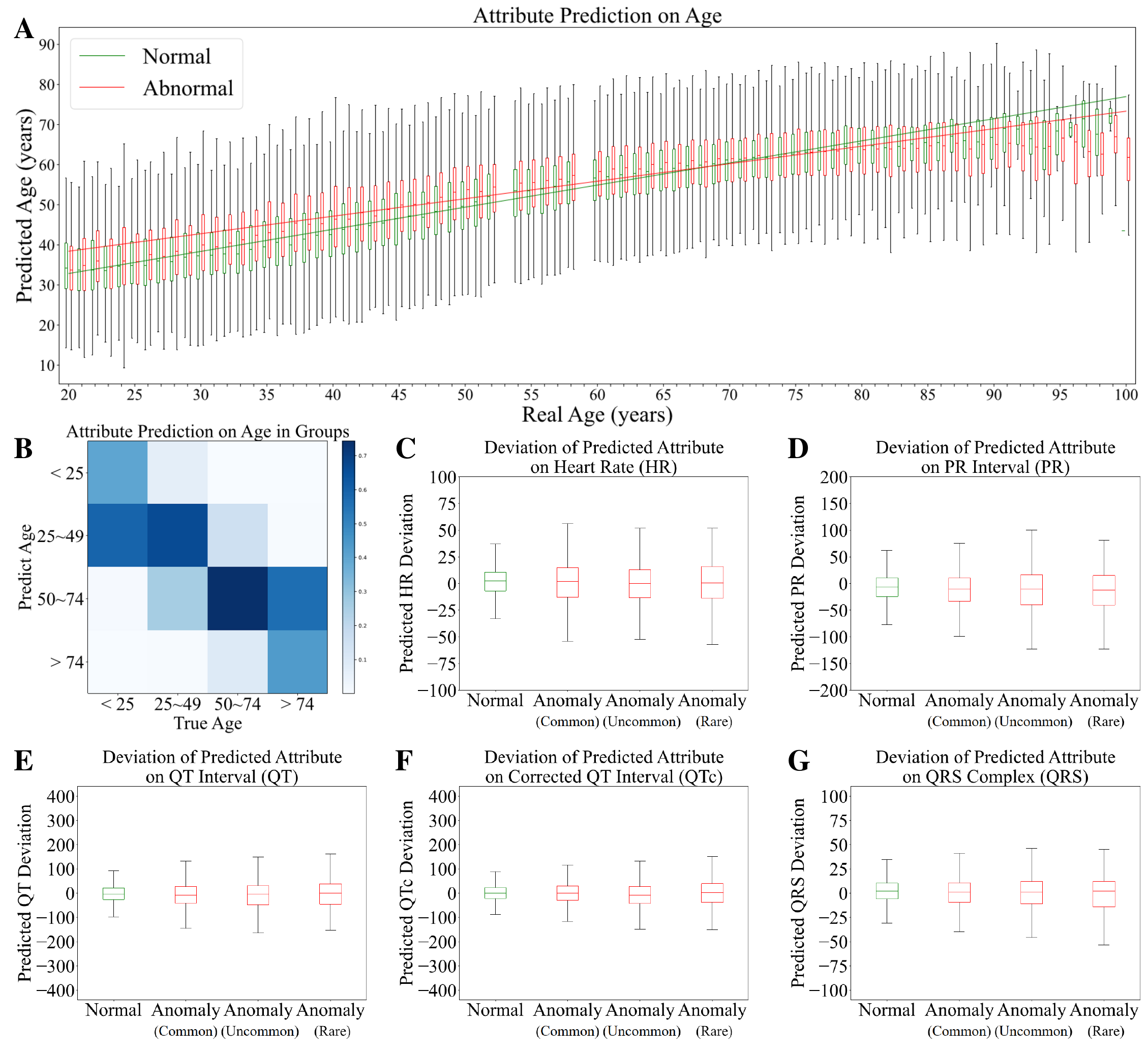}
\caption{Evaluation of the attribute prediction module: A) Comparison of predicted ages across different age groups for normal and abnormal ECGs; B) Age classification accuracy for normal ECGs depicted across age ranges, showing predicted versus actual age; C) Heart rate prediction deviation compared to a standard reference range (y-axis) across varying anomaly rarity (x-axis); D-G) Analysis of deviations in PR interval, QT interval, corrected QT interval, and QRS complex predictions.}
\label{fig:abl_apm}
\end{figure}

\section{Additional Results Tables and Figures}

The detailed numerical results corresponding to Figure 3 in the main paper are provided in Supplementary Table~\ref{tab:ruijin_detail}. Upon removal of all modules, our method transforms into a simplified reconstruction approach focused solely on the global ECG, utilizing a basic L2 loss function. Using the AUROC metric as a benchmark, we observed significant performance enhancements with the introduction of MR and MC modules. For instance, the AUROC for the standard test set increased from 78.1\% to 88.2\% and then to 90.7\% upon integrating MR and MC, respectively. Similar improvements were noted for the uncommon and rare test sets, with the AUROC climbing from 77.1\% to 87.7\% and 88.6\% for the uncommon set, and from 73.8\% to 88.4\% for the rare set. TAR further improved detection in common and uncommon test sets but slightly decreased performance for rare anomalies. Adding the TAR further boosted the detection of standard and uncommon anomalies, achieving 91.1\% AUROC for standard anomalies and 88.8\% for uncommon anomalies. However, it slightly reduced the detection performance for rare anomalies by 0.1\% AUROC. The final incorporation of the APM improved detection across all anomaly types, leading to the highest recorded AUROC scores: 91.4\% for standard, 89.5\% for uncommon, and 89.6\% for rare anomalies. With the step-by-step integration of MR, MC, TAR, and APM modules to the simplified reconstruction model, there is a progressive enhancement in AUROC values and F1 scores on all the test sets, emphasizing the cumulative impact of all modules on the suggested framework.

Additionally, the box plot in Supplementary Figure~\ref{fig:abl_apm} shows that irrespective of whether dealing with normal or anomalous data, our attribute prediction module demonstrates a generally commendable performance. In Supplementary Figure~\ref{fig:abl_apm}-A, the predicted age on normal ECG exhibits a narrower deviation interval and demonstrates a more positive correlation with real age, in contrast to abnormal ECG predictions. The model’s age categorization accuracy for normal ECGs is demonstrated in Supplementary Figure~\ref{fig:abl_apm}-B, where the model successfully categorizes ages into correct groups with an accuracy exceeding 70\%. Notably, as the ECG transitions from a normal to an abnormal state, or from a common anomaly to a rare one, the predictive performance deteriorates, as detailed from Supplementary Figure~\ref{fig:abl_apm}-C to Supplementary Figure~\ref{fig:abl_apm}-G, indicating a positive correlation between the degree of rarity and prediction accuracy.

\begin{table}[t]
\centering
\caption{Anomaly detection and anomaly localization results on PTB-XL database. Results are shown in the patient-level AUC for anomaly detection and the signal point-level AUC for anomaly localization, respectively. The best-performing method is in \textbf{bold}, and the second-best is \underline{underlined}.}
\scalebox{0.85}{
\begin{tabular}{C{3.5cm}|C{1.7cm}C{1.9cm}C{1.4cm}C{1.6cm}C{1.6cm}C{1.1cm}C{2.0cm}|C{1.0cm}} 
\hline
Metrics & DAGMM~\cite{zong2018deep} & MADGAN~\cite{li2019mad} & USAD~\cite{audibert2020usad} & TranAD~\cite{tranad} & AnoTran~\cite{xu2022anomaly} & TSL~\cite{zheng2022task} & BeatGAN~\cite{liu2022time} & Ours\\
\hline
patient-level AUC & 0.782 & 0.775 & 0.785 & 0.788 & 0.762 & 0.757 & \underline{0.799} & \textbf{0.860}\\
signal point-level AUC & 0.688 & 0.708 & 0.683 & 0.685 & 0.641 & 0.509 & \underline{0.715} & \textbf{0.747}\\
\hline
\end{tabular}}
\label{tal:ptxb}
\end{table}

\begin{table}[t]
\centering
\caption{Anomaly detection results on MIT-BIH dataset. Results are shown in terms of the AUC and F1 score for heartbeat-level classification. The best-performing method is in \textbf{bold}, and the second-best is \underline{underlined}.}
\scalebox{0.85}{
\begin{tabular}{C{3.5cm}|C{1.7cm}C{1.8cm}C{1.4cm}C{1.6cm}C{1.6cm}C{1.1cm}C{2.0cm}|C{1.0cm}} 
\hline
Metrics & DAGMM~\cite{zong2018deep} & MSCRED~\cite{zhang2019deep} & USAD~\cite{audibert2020usad} & TranAD~\cite{tranad} & AnoTran~\cite{xu2022anomaly} & TSL~\cite{zheng2022task} & BeatGAN~\cite{liu2022time} & Ours\\
\hline
heartbeat-level AUC & 0.700 & 0.627 & 0.352 & 0.742 & 0.770 & 0.894 & \underline{0.945} & \textbf{0.969}\\
F1 Score & 0.677 & 0.778 & 0.384 & 0.621 & 0.650 & 0.750 & \underline{0.816} & \textbf{0.883}\\
\hline
\end{tabular}}
\label{tal:mit}
\end{table}

\section{Additional Experiments on Publicly Available ECG Datasets}

\subsection{Results on PTB-XL Dataset}

The performance of anomaly detection on PTB-XL is succinctly presented in Supplementary Table~\ref{tal:ptxb}. The proposed method demonstrates notable efficacy with an 86.0\% AUC in patient-level anomaly detection, surpassing all baseline models by a considerable margin (10.3\%). It also outlines the outcomes of anomaly localization across diverse individuals using our proposed benchmark.

At the signal point level, the proposed method achieves an AUC of 74.7\%, outperforming all baseline models (3.2\% higher than BeatGAN). It is imperative to highlight that TSL, originally not designed for localization, exhibits suboptimal performance in this specific task.

\subsection{Results on MIT-BIH Dataset}

Supplementary Table~\ref{tal:mit} presents the comparative outcomes on MIT-BIH, showcasing the performance of the proposed method. The proposed approach attains a heartbeat-level AUC of 96.9\%, reflecting a notable enhancement of 2.4\% compared to the state-of-the-art BeatGAN, which achieves an AUC of 94.5\%. Additionally, the F1-score for the proposed method is 88.3\%, exhibiting a substantial improvement of 6.7\% over BeatGAN's F1-score of 81.6\%.

\end{refsection}
\end{document}